\documentclass{article} 

\usepackage{iclr2017_conference,times}
\usepackage{url}

\usepackage{epsfig}
\usepackage{graphicx}
\usepackage{amsmath}
\usepackage{amssymb}
\usepackage{array}
\usepackage{textcomp}
\usepackage{multirow}
\usepackage{subcaption}
\graphicspath{{./images/}}
\usepackage[toc,title]{appendix} 
\usepackage{tabulary}

\usepackage{hyperref}

\title{What Is the Best Practice for CNNs Applied to Visual Instance Retrieval?}

\author{Jiedong Hao, Jing Dong, Wei Wang, Tieniu Tan \\
Center for Research on Intelligent Perception and Computing\\
Institute of Automation, Chinese Academy of Sciences\\
}

\begin{document}
\maketitle

\begin{abstract}
Previous work has shown that feature maps of deep convolutional neural networks (CNNs)
can be interpreted as feature representation of a particular image region. Features aggregated from
these feature maps have been exploited for image retrieval tasks and achieved state-of-the-art performances in
recent years. The key to the success of such methods is the feature representation. However, the different
factors that impact the effectiveness of features are still not explored thoroughly. There are much less
discussion about the best combination of them.

The main contribution of our paper is the thorough evaluations of the various factors that affect the
discriminative ability of the features extracted from CNNs. Based on the evaluation results, we also identify
the best choices for different factors and propose a new multi-scale image feature representation method to
encode the image effectively. Finally, we show that the proposed method generalises well and outperforms
the state-of-the-art methods on four typical datasets used for visual instance retrieval.
\end{abstract}

\section{Introduction}

Image retrieval is an important problem both for academic research and for industrial applications. Although it
has been studied for many years~\citep{sivic2003video,Philbin07,Tolias2016}, it is still a challenging task.
Generally, image retrieval is divided into two groups. The first one is the category-level image
retrieval~\citep{sharma2015}, in which an image in the dataset is deemed to be similar to the query
image if they share the same class or they are similar in shape and local structures.
The other group is the instance-level image retrieval~\citep{Tolias2016}, in which
an image is considered to match the query if they contain the same object or the same scene. The instance-level
image retrieval is harder in that the retrieval method need to encode the local and detailed information in
order to tell two images apart, \textit{e.g.}, the algorithm should be able to detect the differences between
the Eiffel Tower and other steel towers although they have similar shapes.
In this paper, we focus on the instance-level image retrieval.

Traditionally, visual instance retrieval is mainly addressed  by the BoF (bag of features) based methods using
the local feature descriptors such as SIFT~\citep{Lowe2004}.
In order to boost the retrieval performances, post-processing
techniques such as query expansion~\citep{chum2007total} and spatial verification \citep{Philbin07} are also
employed.

With the decisive victory~\citep{krizhevsky2012} over traditional models in the ImageNet~\citep{ILSVRC15} image
classification challenge, convolutional neural networks~\citep{LeCun98} continue to achieve remarkable
success in diverse fields such as object
detection~\citep{liu15ssd,ren15fasterrcnn}, semantic segmentation~\citep{dai2016instance} and even image style
transfer~\citep{Gatys_2016_CVPR}.
Networks trained on the Imagenet classification task can generalize quite well to other tasks,
which are either used off-the-shelf~\citep{Razavian2014a} or fine-tuned on the task-specific datasets~\citep{AzizpourRSMC14,
long2015fully}. Inspired by all these, researchers in the field of image retrieval also shift their
interest to the CNNs. Their experiments have shown promising and surprising results~\citep{Babenko2014,Razavian2014b,
Tolias2016}, which are on par with or surpass the performances of conventional
methods like BoF and VLAD (vector of locally aggregated descriptors)~\citep{Jegou2010cvpr, Relja2013} .

Despite all these previous advances~\citep{Babenko2014, Babenko2015, Tolias2016} on using CNNs for image
feature representation, the underlying factors that contribute to the success of off-the-shelf CNNs on the
image retrieval tasks are still largely
unclear and unexplored, \textit{e.g., which layer is the best choice for instance retrieval, the convolutional
layer or the fully-connected layer? What is the best way to represent the multi-scale information of an image?}
Clarifying these questions will help us advance a further step towards building a more robust and accurate
retrieval system. Also in situations where a large numbers of training samples are not available, instance
retrieval using unsupervised method is still preferable and may be the only option.

In this paper, we aim to answer these questions and make three novel contributions. Unlike previous
papers, we explicitly choose five factors to study the image representations based on CNNs
and conduct extensive experiments to evaluate their impacts on the retrieval performances.
We also give detailed analysis on these factors and give our recommendations for combining them.
During experiments, we borrow wisdoms from literatures and evaluate their usefulness, but find that
they are not as effective as some of the simpler design choices.
Second, by combining the insights obtained during the individual experiments, we are able to propose a
new multi-scale image representation, which is compact yet effective. Finally, we evaluate our method on
four challenging datasets, \textit{i.e.}, Oxford5k, Paris6k, Oxford105k and UKB.
Experimental results show that our method is generally applicable and outperforms all previous methods
on compact image representations by a large margin.

\section{Related Work}

\noindent
\textbf{Multi-scale image representation}.
\citet{Lazebnik2006} propose the spatial pyramid matching approach to encode the spatial information using
BoF based methods. They represent an image using a pyramid of several levels or scales. Features
from different scales are combined to form the image representation in such a way that coarser levels get less
 weight while finer levels get more weight. Their argument is that matches found in coarser levels
may involve increasingly dissimilar image features. In our paper, we also explore the multi-scale paradigm in the
same spirit using the convolutional feature maps as the local descriptors. We find that the deep features from
the convolutional feature maps are distinct from the traditional descriptors: the weighted sum of different
level of features shows no superior performances than a simple summation of them.
\citet{kaiming14ECCV} devise an approach called SPP (spatial pyramid pooling).
In SPP, feature maps of the last convolutional layer are divided into a 3 or 4 scale pyramid. First the regional
features in each scale are concatenated, then the scale-level features are concatenated to a fixed length vector
to be forwarded to the next fully-connected layers. We find that this
strategy is ineffective for unsupervised instance retrieval, leading to inferior performances compared to
other simple combination methods (see the part about multi-scale representation in section~\ref{subsec5.2}
for more details.).

\noindent
\textbf{Image representation using off-the-shelf CNNs}. \citet{Gong2014} propose the MOP (multi-scale
orderless pooling) method to represent an image in which VLAD is used to encode the level 2 and level 3
features. Then features from different scales are PCA-compressed and concatenated to form the image features.
This method is rather complicated and time-consuming. At the same time, \citet{Babenko2014} use
Alexnet~\citep{krizhevsky2012} trained on the Imagenet 1000-class classification task and retrain the
network on task-related dataset. The retraining procedure gives a boost to the retrieval performances.
Instead of using the output of the fully-connected layers as the image feature representations,~\citet{Babenko2015}
use the output feature maps of last convolutional layer to compute the image features.
Recently, instead of sum-pooling the convolutional features, \citet{Tolias2016} use max-pooling
to aggregate the deep descriptors. Their multi-scale method, called R-MAC (regional maximum activation of
convolutions), further improves the previous results on four common instance retrieval datasets.
Our work differs from these papers in that we explicitly explore the various factors that underpin
the success of unsupervised instance retrieval, which have not been fully explored and analysed.
By carefully choosing the different setting for each factor and combining them in a complementary way,
we show that a large improvement can be achieved without additional cost.
\section{Impacting Factors}

When we employ off-the-shelf CNNs for the task of instance-level image retrieval, a natural question
is: what kind of design choices should we make in order to make full use of the representational power of
existing models? In this section,
we summarize the five factors that may greatly impact the performance of the final image retrieval
system. In section~\ref{subsec5.2}, we will show our experimental results on each key factor. Before we
delve into the impacting factors, first we will give a brief introduction about how to represent an image
using the activation feature maps of a certain layer.

\subsection{CNN Features for Instance Retrieval}
In this paper, we are mainly interested in extracting compact and discriminative image features using the
off-the-shelf CNNs in an efficient way.
For a given image $I$, we simply subtract the mean value of the RGB channels from the original image
and do not do other sophisticated preprocessing. Then the image is fed into the convolutional network
and goes through a series of convolutions, non-linear activations and pooling operations.
The feature activation maps of a certain layer can be interpreted as the raw image
features, based on which we build the final image features. These feature maps form a tensor of
size $K\times H \times W$, where
$K$ is the number of feature channels, and $H$ and $W$ are height and width of a feature map.
Each feature map represents a specific pattern which encodes a small part of information about the
original image.
If we represent the set of feature maps as $F=\{F_i\}, i=1,2,\ldots, K$, where $F_i$ is the $i^{th}$
activation feature map, then the most simple image feature is formulated as:
\begin{equation}\label{eq:single_scale}
	f = [f_1, f_2, \ldots, f_i, \ldots, f_K]^T.
\end{equation}
In the above equation~\ref{eq:single_scale}, $f_i$ is obtained by applying the feature aggregation
method (see section~\ref{subsec3.2}) over the $i^{th}$ feature map $F_i$. Throughout this paper, we use feature
maps after the non-linear activations (ReLU) so that the elements in each feature map are all
non-negative. We also experiment with feature maps prior to ReLU, but find that they
lead to inferior performances. After the image feature representation is obtained, post-processing
techniques such as PCA and whitening can be further applied.

\subsection{Impacting Factors on Performance}\label{subsec3.2}
\noindent
\textbf{Feature aggregation and normalization.} After the feature maps of a certain layer are
obtained, it is still challenging to aggregate the $3$-dimensional feature maps to get compact vector
representations for images. Previous papers use either sum-pooling~\citep{Babenko2015} or max-pooling~
\citep{Tolias2016} followed by $l_2$-normalization.
Sum-pooling over a particular feature map $F_i$ is expressed as
\begin{equation}
	f_i = \sum_{m=1}^{H}\sum_{n=1}^{W}F_i(m, n), i\in\{1, 2, \ldots, K\},
\end{equation}
while max-pooling is given by
\begin{equation}
	f_i = \max_{m,n} F_i(m,n),
\end{equation}
where $m, n$ are all the possible values over the spatial coordinate of size $H\times W$.
In this paper, for the first time, different combinations of aggregation and normalization
methods~($l_2$ and $l_1$ in the manner of RootSIFT~\citep{Relja2012}) are evaluated and their results
are reported.

\noindent
\textbf{Output layer selection.} \citet{zeiler2014} has shown that image
features aggregated from the feature activation maps of certain layers have interpretable semantic
meanings. \citet{Gong2014} and \citet{Babenko2014} use the output of the
first fully-connected layer to obtain the image features, while~\cite{Babenko2015} and~\cite{Tolias2016}
use the output feature maps of the last convolutional layer. But these choices are somewhat
subjective. In this paper, we extract dataset image features from the output feature maps of different
layers and compare their retrieval performances. Based on the finding in this experiment, we choose the
best-performing layer and also come up with a layer ensemble approach which outperforms state-of-the-art
methods (see section~\ref{subsec:comparision}).

\noindent
\textbf{Image resizing.} Famous models such as Alexnet~\citep{krizhevsky2012} and VGGnet~\citep{Simonyan14}
all require that the input images have fixed size. In order to meet this requirement, previous
 papers~\citep{Gong2014, Babenko2015} usually resize the input images to the fixed size. We postulate that
 the resizing operation may lead to the distortion of important information about the objects in the
natural images. Ultimately, this kind of operation may hurt the discriminative power
of image features extracted from the network, thus degrading the retrieval performances. For the
task of image retrieval, we think it is best to keep the images their original sizes and feed them
directly to the network whenever possible.
In this paper, three image resizing strategies are explored:
 \begin{itemize}
	 \item Both the height and width of the dataset images are set to the same fixed value (denoted as
	  \textit{two-fixed}).
	 \item The minimum of each dataset image's size is set to a fixed value. (The aspect ratio of
    the original image is kept.) (denoted as \textit{one-fixed}).
	 \item  The images are kept their original sizes. (denoted as \textit{free}).
 \end{itemize}

\noindent
\textbf{Multi-scale feature representation.} Unlike local feature descriptors such as SIFT~\citep{Lowe2004},
the feature vector extracted from the deep convolutional networks for an image is a global descriptor
which encodes the holistic information. When used for image retrieval, this kind of
features still lack the detailed and local information desired to accurately match two images. Inspired
by spatial pyramid matching~\citep{Lazebnik2006} and SPP~\citep{kaiming14ECCV}, we explore the feasibility
of applying this powerful method to obtain discriminative image features. An image is represented
by a $L$-level pyramid, and at each level, the image is divided evenly into several overlapping or non-overlapping
regions. The vector representations of these small regions are
computed, then the regional vectors are combined to form the image feature vectors.
The single scale representation of an image is just a special case of the multi-scale method in which
the number of level $L$ equals 1.

\begin{figure}[t]
	\centering
	\begin{subfigure}{.25\columnwidth}
		\centering
		\includegraphics[width=\linewidth]{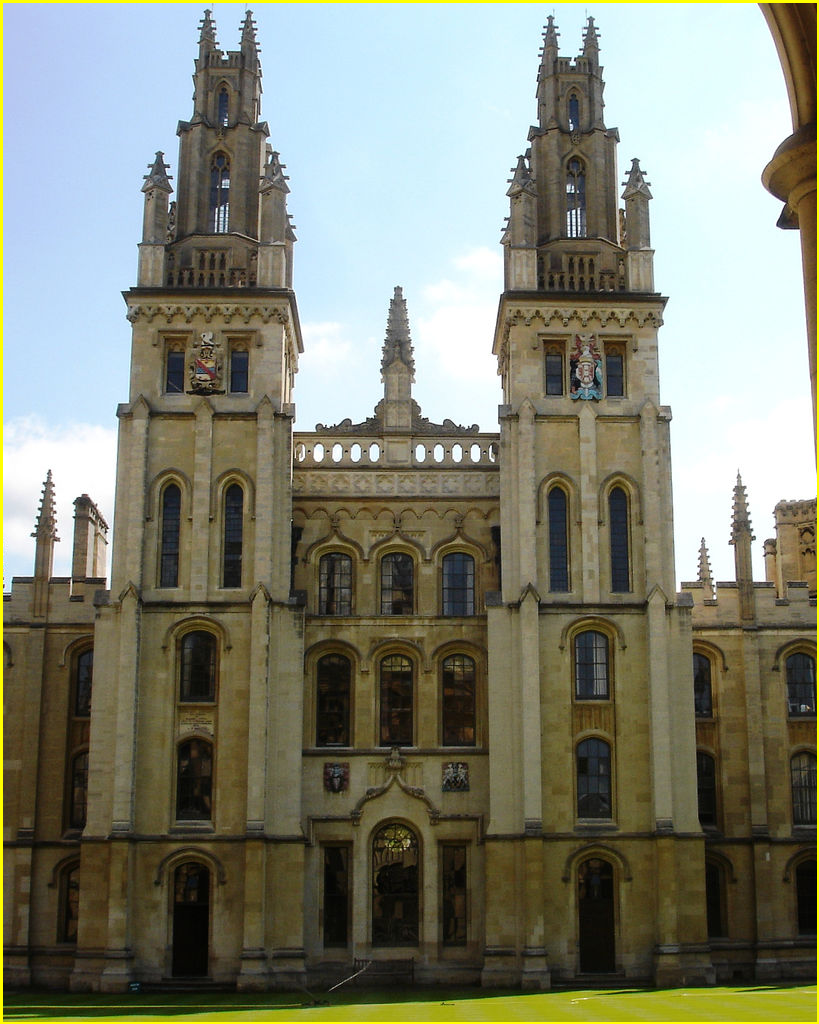}
		\subcaption{level 1}
	\end{subfigure}%
	\hfill
	\begin{subfigure}{.25\columnwidth}
		\centering
		\includegraphics[width=\linewidth]{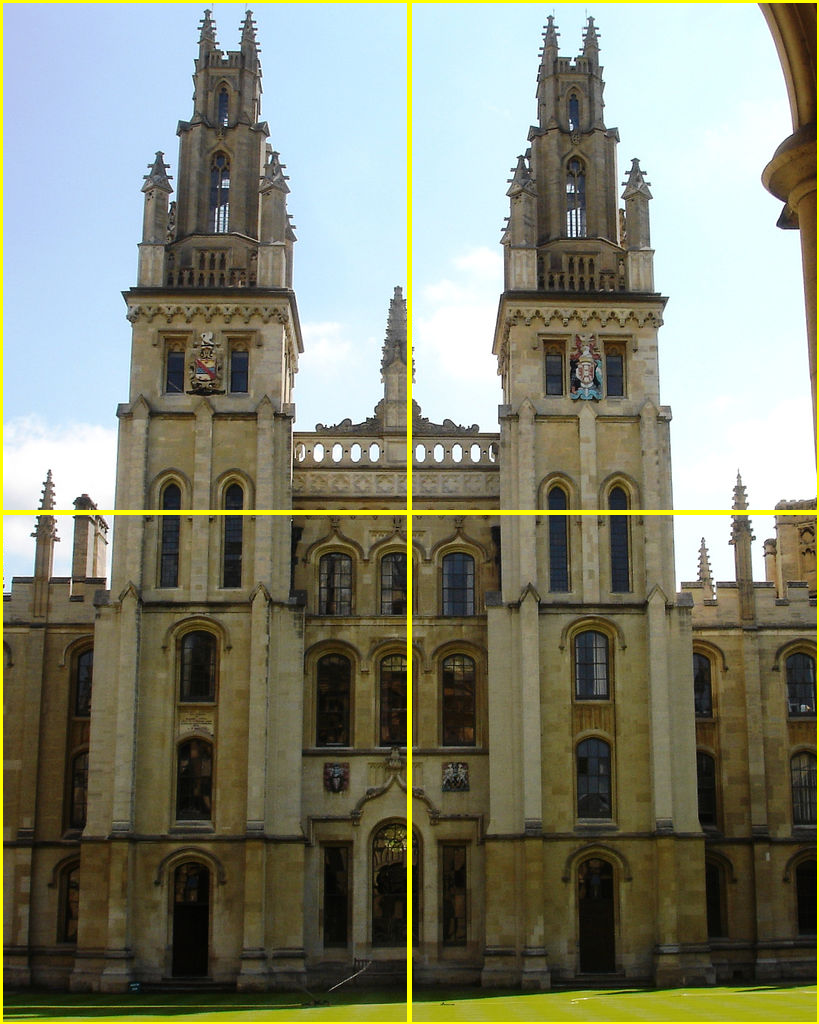}
		\subcaption{level 2}
	\end{subfigure}%
	\hfill
	\begin{subfigure}{.25\columnwidth}
		\centering
		\includegraphics[width=\linewidth]{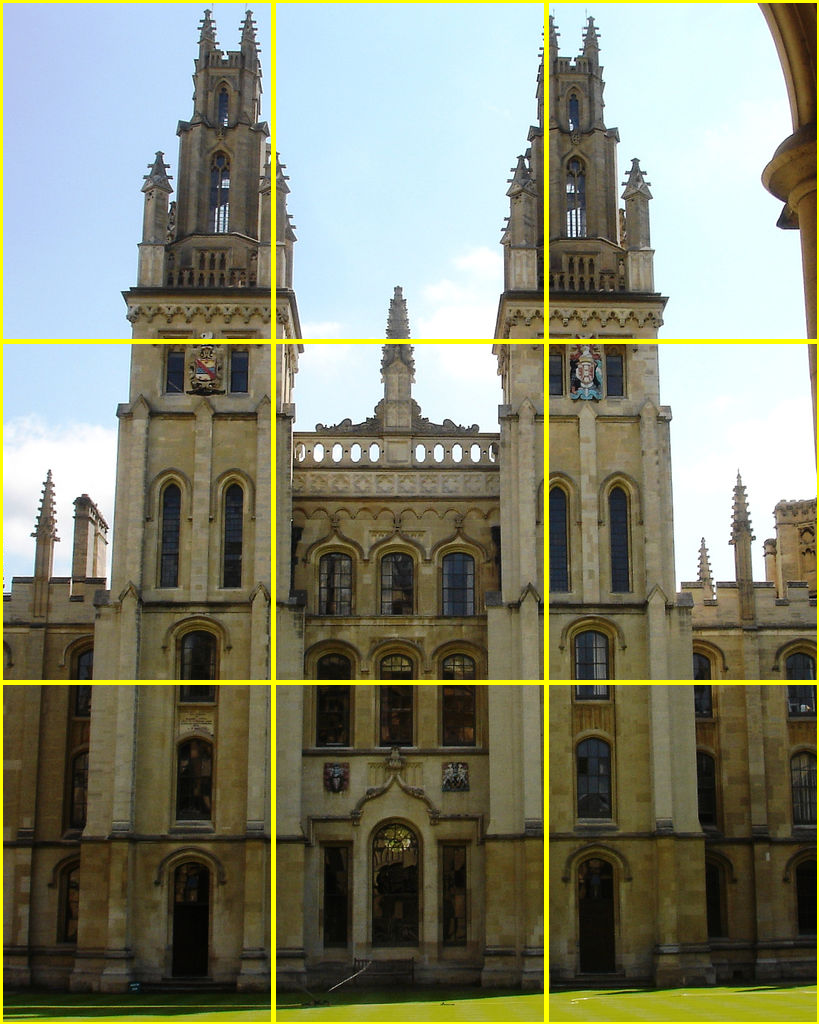}
		\subcaption{level 3}
	\end{subfigure}
	\caption{{\small \textbf{An illustration of multi-scale representation of an image.} The whole
			image is divided into 3 levels from the coarsest (level 1) to the finest (level 3).
			At each level, the image is divided into different number of equal-sized regions.}}
	\label{fig:multiscale}
\end{figure}

Figure~\ref{fig:multiscale} shows
an example of $3$ level representations of an image. The time cost of re-feeding those small regions
into the network to compute the regional vectors would be huge, thus unacceptable for instance retrieval
tasks. Inspired by the work of \citet{girshick2015_fast_rcnn} and \citet{Tolias2016},
we assume a linear projection between the original image regions and the regions in the feature maps of a
certain layer. Then the regional feature vectors can be efficiently computed without re-feeding the
corresponding image regions. In section~\ref{subsec5.2}, various settings for the multi-scale and
scale-level feature combination methods are explored and their retrieval performances are
reported and analysed.

\noindent
\textbf{PCA and whitening.} Principal Component Analysis (PCA) is a simple yet efficient method for
reducing the dimensionality of feature vectors and decorrelating the feature elements. Previous
work~\citep{Babenko2014,
Jegou2010cvpr} has shown evidences that PCA and whitened features can actually boost the
performances of image retrieval. In this paper, we further investigate the usefulness of PCA and
whitening within our pipeline and give some recommendations.

\section{Implementation}
We use the open source deep learning framework Caffe~\citep{jia2014caffe} for our whole experiments.
The aim of this research is to investigate the most effective ways to exploit the feature activations
of existing deep convolutional models. Based on past practices for networks to go
deeper~\citep{krizhevsky2012,Simonyan14,Szegedy2015, He2015}, a consideration for moderate
computational cost, and also the results from~\citet{Tolias2016} that deeper networks work better than
 shallower ones, we decide to use the popular VGG-19 model~\citep{Simonyan14} trained on ImageNet as
 our model.

\noindent
\textbf{Network transformation}.
The original VGG-19 network only accepts an image of fixed size ($224\times 224$),
which is not the optimal choice when extracting image features for retrieval tasks.
In order for the network to be able to process an image of arbitrary size (of course,
the image size can not exceed the GPU's memory limit) and for us to experiment with different
input image resizing strategies, we adapt the original VGG-19 network and change the
fully-connected layers to convolutional~\citep{long2015fully} layers. For more details about
network transformations, see appendix~\ref{app:trans}.


\section{Experiments}
In this section,  we first introduce the datasets used and the evaluation metrics. Then we report
our experimental results for different impacting factors and give detailed analysis. In the last
part, we show the performance of our method considering all these impacting factors and compare
our method with the state-of-the-art methods on four datasets.

\subsection{Datasets and Evaluation Metrics}

The \textbf{Oxford5k} dataset~\citep{Philbin07} contains 5062 images crawled from
Flickr by using 11 Oxford landmarks as queries. A total of 11 groups of queries --- each having 5
queries with their ground truth relevant image list, are provided. For each query, a bounding box
annotation is also provided to denote the query region. During experiment, we report results using
the full query images (denoted as full-query) and image regions within the bounding boxes of the query
images (denoted as cropped-query). The performance on this dataset is measured by mAP (mean average precision)
over all queries.

The \textbf{Paris6k} dataset~\citep{Philbin08} includes 6412 images\footnote{
Following conventions, 20 corrupted images from this dataset are removed, leaving 6392 valid images.}
from Flickr which contains 11 landmark buildings and the general scenes from Paris. Similar to
the Oxford5k dataset, a total of 55 queries belonging to 11 groups and the ground truth bounding boxes
for each query are provided . The performance is reported as mAP over 55 queries.

The \textbf{Oxford105k\footnote{The image named
 ``portrait\_000801.jpg'' was corrupted and manually removed from this dataset.}}
dataset contains the original Oxford5k
dataset and additional 100,000 images~\citep{Philbin07} from Flickr. The 100,000 images are disjoint with
the Oxford5k  dataset and are used as distractors to test the retrieval performance when
the dataset scales to larger size. We use the same evaluation protocol as the Oxford5k
on this dataset.

The \textbf{UKB} dataset~\citep{nister2006cvpr} consists of
10200 photographs of 2550 objects, each object having exactly 4 images. The pictures of these
objects are all taken indoor with large variation in orientation, scale, lighting and shooting angles.
During experiment, each image is used to query the whole dataset. The performance is measured by the
average number of same-object images in the top-4 results.

\subsection{Results and Discussion}	\label{subsec5.2}
In this section, we report the results of experiments on the impact of different factors and analyse their
particular impact. The experiments in this section are conducted on the Oxford5k dataset.

\begin{table}[t]
\centering
\begin{minipage}[b]{0.46\columnwidth}
  \centering
  \small
  \caption{{\small \textbf{Comparison between different combinations of feature aggregation and
			normalization methods.} }}
  \label{table:aggre_norm}
		\begin{tabular}{|c|c|c|}
			\hline
			Method	&	full-query & cropped-query \\
			\hline
			$max$-$l_1$  &	52.4	  &	48.0 \\
			$sum$-$l_2$  &	58.0 	 &  52.6	\\
			$sum$-$l_1$  &    60.3	&	56.3   \\
			$max$-$l_2$	&	60.1	 &  53.5  \\
			\hline
		\end{tabular}
\end{minipage}
\hfill
\begin{minipage}[b]{0.46\columnwidth}
\centering
\small
\caption{{\small \textbf{Comparison between different image resizing strategies.}
			 The numbers in the parentheses denote
			the sizes in which the maximum mAPs are achieved.
            }}		
	\label{table:img_size}
    \begin{tabular}{|c|c|c|}
		\hline
		Method		    & full-query			& cropped-query \\
		\hline
		\textit{two-fixed}		&		55.5 (864)		& 	38.7 (896)	\\
		\textit{one-fixed}		&		59.0 (800) 		&	39.3 (737)\\
		\textit{free}			&		58.0	 		&	52.6 	\\
		\hline
	\end{tabular}
\end{minipage}
\end{table}
\noindent
\textbf{Feature aggregation and normalization.} In this experiment, we compare the different combinations of
 feature aggregation (sum-pooling and max-pooling) and normalization methods ($l_2$ and $l_1$) in terms of
 their retrieval performances. We use features from the layer conv5\_4 with the \textit{free} input image
 size. The results (\%) are shown in Table~\ref{table:aggre_norm}.
Sum-pooling followed by $l_1$ normalization leads to slightly better results than the other combinations,
especially for the cropped-query. However, after preliminary experiment with a multi-scale
version of $sum$-$l_1$ and $max$-$l_2$, we find that $max$-$l_2$ is much better than $sum$-$l_1$.
For example, employing a 4 level representation of images in the Oxford5k dataset,
for the case of full-query, we find that the mAP for the $max$-$l_2$ method is 65.1, while the mAP for
$sum$-$l_1$  is only 51.3 (even lower than the single scale representation). Base on these results,
we stick to $max$-$l_2$ in computing the final image features.

\noindent
\textbf{Output layer selection.} In order to verify their feasibility for instance retrieval, we extract
from the network the output feature maps of different layers and aggregate them to get the image
feature vectors. We evaluate the performances using features from layer
conv3\_3 up to the highest fc7-conv layer (except the pooling layers, \textit{i.e.} pool3, pool4 and pool5).
Single-scale representations of the dataset images are used in this experiment.

\begin{figure}[htpb]
	\centering
		\includegraphics[width=\columnwidth]{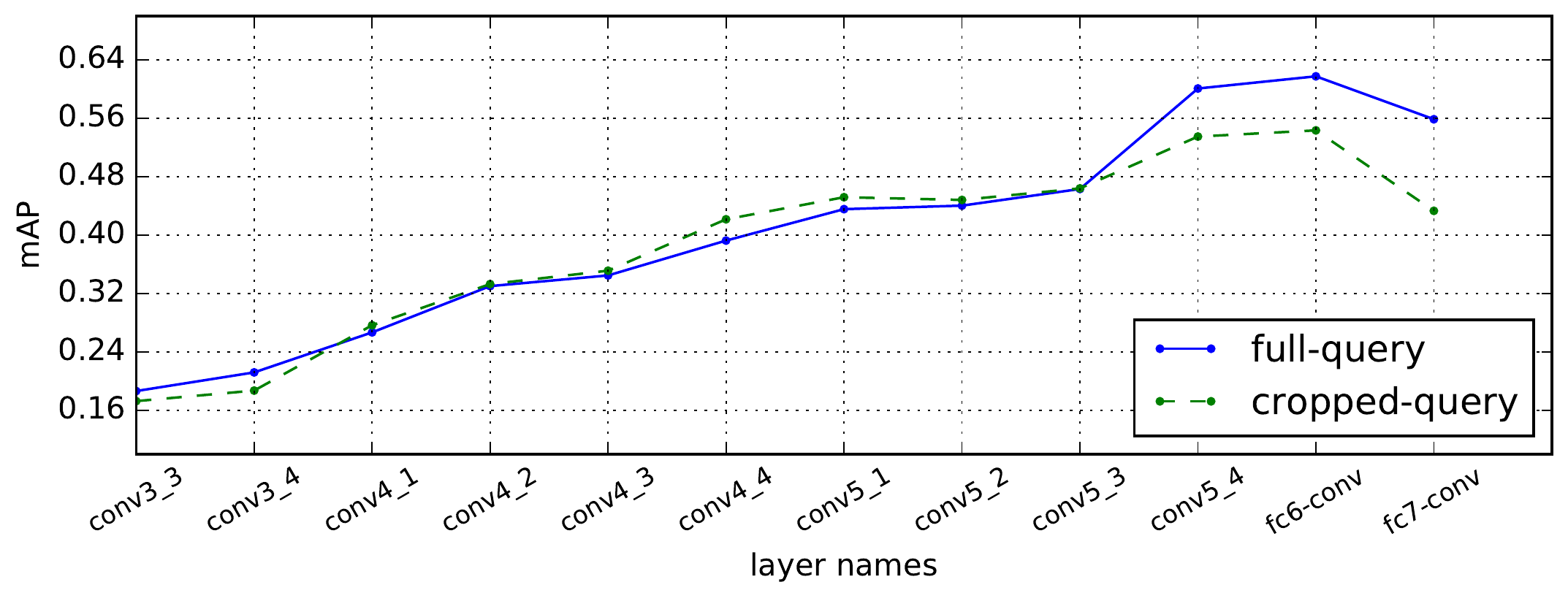}
	\caption{{\small
			 \textbf{Performance comparison between different layers.} This
			experiment is conducted using the \textit{free} input image size.}}
	\label{fig:different_layer}
\end{figure}

Figure~\ref{fig:different_layer} shows the retrieval performances of image features corresponding to
different layers. The retrieval performances for both the full and cropped queries increase as the
layer increases from lower layer conv3\_3 to higher layers and plateau in layer conv5\_4 and fc6-conv,
then the performances begin to decrease as the layers increase to fc7-conv. The result shows that
features from lower layers such as conv3\_3 and conv3\_4 are too generic and lack the
semantic meanings of the object in the image, thus rendering them unsuitable for instance retrieval. On the
other hand, features from the highest layer (fc7-conv) contain the semantic meaning of objects but
lack the detailed and local information needed to match two similar images. The best results are
obtained in layer conv5\_4 (0.601) and fc6-conv (0.618), where the feature vectors combine both the
low-level detailed information and high level semantic meanings of the image.
Based on these observations and the requirement for keeping the image
features compact, we mainly focus on image features from the layer conv5\_4 (dimensionality = 512
 compared to 4096 of layer fc6-conv).

\noindent
\textbf{Image resizing.} We experiment with 3 kinds of image resizing strategies which are detailed in
section~\ref{subsec3.2}.
We use grid search to find the optimal size for the \textit{two-fixed} and \textit{one-fixed}
strategy. As is shown in Table~\ref{table:img_size}, the \textit{free} input strategy outperforms or is close to
the other two strategies: it performs especially well in the cropped-query case. This
experiment shows that changing the image aspect ratio (\textit{two-fixed}) distorts the image information,
thus reducing the performance dramatically. The \textit{one-fixed} way is better
than the \textit{two-fixed} method. But information loss still occurs due to the resizing
operation. The \textit{free} method is able to capture more natural and un-distorted
information from the images, which explains its superior performance over the other two methods.
It is best to keep the images their original sizes for the instance retrieval tasks.

\noindent
\textbf{The benefit of multi-scale representation.} In our multi-scale approach, the regional
vectors from each scale are simply added together and $l_2$-normalized to form the scale-level
feature vectors. Then features from different scales are combined and $l_2$-normalized to form the image
representations. In fact, we also experimented with two methods which concatenate features from different
scales.
The first method is in same vein to spatial pyramid pooling~\citep{kaiming14ECCV}, \textit{i.e.},
region-level as well as the scale-level features are all concatenated to form a high dimensional vector.
In the second method, region-level features are added while scale-level features are concatenated. We find
that these two methods all lead to inferior results. The performance drop for the first in the case of
cropped-query can be as large as 41\%. The high dimensionality of the concatenated features (larger than 1.5k)
will also lead to longer running times. Considering all these, we do not use concatenation of features
in the following experiments.

\begin{table}[h]
	\begin{center}
	\small
    \caption{{\small \textbf{Multi-scale representation: comparison between different methods.}
			``overlap'' denotes whether the regions in each level (see Figure~\ref{fig:multiscale}) have
			some overlapping areas. ``s2'',``s3'' mean that overlap occurs in level 2 or 3.
             ``weighing'' means if the features from each level are added using same weight or
              different weight. ``version'' means the different choice of the number of regions
              in each scale.}}
	\label{table:multiscale}
	\begin{tabular}{|c|c|c|c|c|c|c|}
		\hline
		     & scale &  overlap   &  weighing  & version &     full-query     &    cropped-query    \\ \hline
		(a1) &   2   & \texttimes & \texttimes &    -    &     63.5      &     59.0      \\
		(a2) &   2   & \texttimes & $\checkmark$ &    -    &     63.9      &     61.0      \\ \hline
		(b1) &   3   & \texttimes & \texttimes &    -    &     64.2      &     60.9      \\
		(b2) &   3   & \texttimes & $\checkmark$ &    -    &     62.6      &     61.0      \\
		(b3) &   3   &     s2     & \texttimes &    -    &     64.8      &     60.8      \\ \hline
		(c1) &   4   &     s3     & \texttimes &   v1    &     65.1      &     61.4      \\
		(c2) &   4   &     s3     & $\checkmark$ &   v1    &     64.8      &     60.7      \\
		(c3) &   4   &   s2,s3    & \texttimes &   v1    &     65.5      &     60.8      \\
		(c4) &   4   &   s2,s3    & \texttimes&   v2    &     65.9      &     61.5      \\
		(c5) &   4   &   s2,s3    & $\checkmark$ &   v2    &     65.4      &     61.2      \\
		(c6) &   4   & \texttimes & \texttimes &   v3    &     64.5      &     61.3      \\
		(c7) &   4   &     s3     & \texttimes&   v3    &     65.8      &     62.2      \\
		(c8) &   4   &   s2,s3    & \texttimes&   v3    & \textbf{66.3} & \textbf{62.6} \\ \hline
	\end{tabular}

	\end{center}
\end{table}
We conduct extensive experiments to decide the best configurations for the multi-scale approach and
report our results in Table~\ref{table:multiscale}.
First, we explore the impact of the number of scales on the retrieval performances. For the 2 and 3 scale
representations, The region number for each level are \{$1\times1$, $2\times2$  \}, \{$1\times1$, $2\times2$,
$3\times3$\}. For the 4 scale representation, 3 versions are used and they differ in the number of regions in
each scale: for ``v1'', ``v2'', and ``v3'', the number of regions are \{$1\times1$, $2\times2$, $3\times3$, $4\times4$\},
\{$1\times1$, $2\times2$, $3\times3$, $5\times5$\} and \{$1\times1$, $2\times2$, $3\times3$, $6\times6$\}.
Table~\ref{table:multiscale} (a1)(b1)(c6) show the performances of using 2, 3, and 4 scales to represent
the dataset images, respectively. Clearly, more scale levels improve the results and in the case of
cropped-query, increase the performance by an absolute 2\%.

We also conduct experiments to find whether the weighing of different scales leads to improved performance.
The weighing method for features from different scales is similar to the manner of spatial pyramid
matching~\citep{Lazebnik2006} --- features from coarser level
are given less weight while features from the finer levels are given more weight. Suppose the features of different
scales for an $L$ scale representation are $f^1, f^2, \ldots, f^L$, then the image representation $f$ is
expressed as:
\begin{equation}
	f = \frac{1}{2^{L-1}}f^1 + \sum_{i=2}^{L}\frac{1}{2^{L-i+1}}f^i\text{.}
\end{equation}
More details can be found in~\citet{Lazebnik2006}. Comparing the results of row (a1) and (a2), it seems that weighing
different scales leads to better performance. But after more experiments, we find that the weighing method generally
leads to inferior results as the number of scales increase, \textit{e.g.}, compare the results of row pair(b1)(b2)
and (c1)(c2). These results suggest that deep features are different from the traditional local
feature descriptors such as SIFT. We should exercise with caution when we apply the traditional
wisdom found in SIFT to the deep convolutional descriptors, which is also suggested in~\citet{Babenko2015}.
Based on the results of this experiment, no weighing methods are used in computing our
final image feature representations.

Next, we look into the issue of overlapping between different scales and try to verify its usefulness. For each
scale and its different versions, we set some overlapping areas between the neighboring regions in either one or two
scales of the pyramid (For the exact configurations of overlap in all cases in Table~\ref{table:multiscale},
see appendix \ref{app:overlap} for the complete descriptions).
From the row pair (b1)(b3)  and (c1)(c3), we can see that overlap increase the performance for full-query but
decrease a little the performance for cropped-query. But for 4 scale v3 (note the pair(c7)(c8)), we see a
consistent improvement for both the full and cropped queries. So we decided to use overlap in level 2 and
3 in computing our final features.

\begin{figure}[t]
	\begin{center}
		\includegraphics[width=\textwidth]{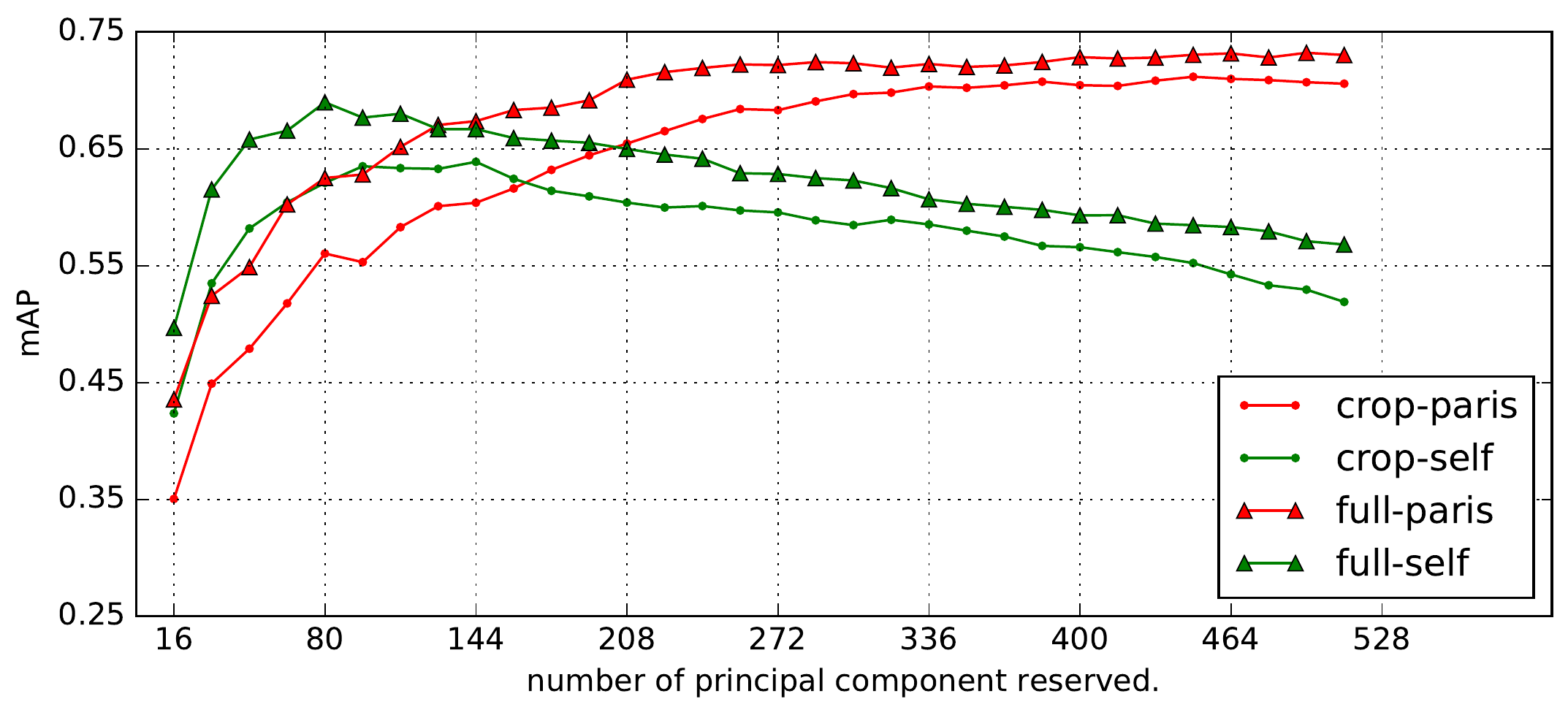}
	\end{center}
	\caption{{\small \textbf{The number of principal component reserved VS mAP.}
			We show the results of full and cropped query using the PCA and whitening matrix
			learned from the Oxford5k itself and Paris6k, denoted as ``full-self'',
			``full-paris'' and  ``crop-self'', ``crop-paris''.}}
	\label{fig:pca_self_paris}
	
\end{figure}

\begin{table}[h]
	\begin{center}
		\small
	
		\caption{{\small \textbf{The impact of PCA and whitening.} ``PCA on self'' and ``PCA on Paris'' mean
                that the corresponding features are post-processed by the PCA and whitening matrices
				learned on the Oxford5k and Paris6k datasets, respectively. The numbers in the parentheses
				indicate the dimensionality of features used for obtaining the corresponding results.}}

		\label{table:pca_whiten}
		\begin{tabulary}{0.9\textwidth}{|L|C|C|}
			\hline
			\multicolumn{1}{|c|}{Feature}                  & full-query & cropped-query \\ \hline
			3scale\_overlap, original         &    64.8    &     60.8      \\
			3scale\_overlap, PCA on self       &    65.4(80)    &     60.9(112)      \\
			3scale\_overlap, PCA on Paris       &    70.6(464)   &     67.3(480)  \\ \hline
			4scale\_v3\_overlap(s3), original     &    65.1   &     61.4      \\
			4scale\_v3\_overlap(s3), PCA on self   &    66.9(80)   &     61.9(96)      \\
			4scale\_v3\_overlap(s3), PCA on Paris   &    72.3(464)   &     70.8(496)     \\ \hline
			4scale\_v3\_overlap(s2,s3),original    &    66.3    &     62.8      \\
			4scale\_v3\_overlap(s2,s3), PCA on self  &    69.0(80)   &     63.9(144)      \\
			4scale\_v3\_overlap(s2,s3), PCA on Paris &    73.2(496)    &     71.2(448)      \\ \hline
		\end{tabulary}
	\end{center}
\end{table}

\noindent
\textbf{PCA and whitening}.
We perform PCA and whitening for the features extracted from the Oxford5k dataset
using the PCA and whitening matrix learned from the Oxford5k or the Paris6k dataset and
$l_2$-normalize these features to get the final image representations.

The retrieval results for 3 groups of features (from Table~\ref{table:multiscale}(b3)(c1)(c8)) are shown
in Table~\ref{table:pca_whiten}. Clearly, PCA and  whitening lead to better performances.
For all 3 groups of features, PCA and whitening on the same dataset lead to insignificant improvement
both in the case of full and cropped query. But after doing PCA and whitening
on the Paris6k dataset, the results for both the full and cropped queries improve greatly.
In fact, the improvement for the case of cropped-query is even more surprising.
For example, for the third feature group, the improvement are
10.4\% and 13.4\% for the full and cropped queries. It should also be noted that as the
the number of principal component reserved increases, the performance for ``PCA on self'' and
``PCA on Paris'' differs greatly.
As is shown in Figure~\ref{fig:pca_self_paris}, the performance for the former peaks at a relatively
low dimension (around 100) and begins to decrease, while for the latter, the performance increases
as the number of principal component gets larger and then plateaus.

Do the above results mean that we should always compute the PCA and whitening matrix from any datasets
other than the query dataset itself? The short answer is \textbf{no}. We find that for UKB,
learning the PCA and whitening matrix on the Oxford5k dataset shows inferior results compared
to learning the PCA and whitening matrix on UKB itself (about 2\% drop in accuracy).
This may be due to the large differences between the images of the two datasets
as the Oxford5k dataset are mainly images of buildings while the images in UKB are mainly small indoor
objects. We therefore recommend learning the PCA and whitening matrix on a similar dataset to achieve good
performances.

\subsection{Comparison with Other Methods} \label{subsec:comparision}

\newcolumntype{A}{ >{\centering\arraybackslash} m{3cm} }
\newcolumntype{B}{ >{\centering\arraybackslash} m{0.5cm} }
\newcolumntype{W}{ >{\centering\arraybackslash} m{0.7cm} }
\begin{table}[t]
    \centering
		\small
        \caption{{\small \textbf{Comparison with state-of-the-art methods.} ``single'' means
        multi-scale features from single layer (conv5\_4)
		are used. ``single, compression'' uses the same features but compresses them to get the best performances.
        ``layer ensemble'' combines the similarity score from layer conv5\_4 and fc6-conv. The dimensionality of
         the combined feature is set to 1024 for compactness considerations. All our methods use PCA and whitening.
		}}
		\label{table:comparsion_with_soa}
		\begin{tabular}{|A|B|W|W|W|W|W|W|W|}
			\hline
			\multirow{2}*{method} &\multirow{2}*{D} &\multicolumn{2}{c|}{Oxford5k}&\multicolumn{2}{c|}{Paris6k} &\multicolumn{2}{c|}{Oxford105k} & \multirow{2}*{UKB} \\ \cline{3-8}
			 &        & full    & \multicolumn{1}{l|}{cropped}   & full  & \multicolumn{1}{l|}{cropped}   & full    & \multicolumn{1}{l|}{cropped}      &  \\ \hline
			{\scriptsize \citet{Jegou2014cvpr} } & 128              & -                & 43.3             & -                & -                & -                & 35.3             & 3.40         \\
			{\tiny \citet{Relja2012}} & 128              & -                & 44.8             & -                & -                & -                & 37.4             & -                  \\
			{\scriptsize \citet{Jegou2014cvpr}}    & 1024             & -                & 56.0             & -                & -                & -                & 50.2             & 3.51          \\
			\hline
			{\scriptsize \citet{Razavian2014}}             & 256              & 53.3             & -                & 67.0             & -                & 48.9             & -                & 3.38     \\
			{\scriptsize \citet{Babenko2014}}               & 512              & 55.7             & -                & -                & -                & 52.2             & -                & 3.56       \\
		
			{\scriptsize\citet{Babenko2015}}        & 256              & 58.9             & 53.1             & -                & -                & 57.8             & 50.1             & 3.65         \\
			{\scriptsize\citet{Relja2016}} & 256  & 62.5 & 63.5 & 72.0 & 73.5 & - & - & -\\
			{\scriptsize \citet{Tolias2016}}                 & 512              & -                & 66.8             & -                & 83.0             & -                & 61.6        & -         \\
			\hline
			{\scriptsize ours (single)}      & 512              & 73.0             & 70.6             & 82.0             & 83.3             & 68.9             & 65.3             & 3.75       \\
			{\scriptsize ours (single, compression)}  & -             & 73.2& 71.2 &  83.0  & 84.0  & 68.9 & 65.8& 3.76\\
			{\scriptsize ours (layer ensemble)}  & 1024				& \textbf{75.6}& \textbf{73.7}  & \textbf{ 85.7}  & \textbf{85.9}  & \textbf{71.6} & \textbf{69.2}& \textbf{3.81} \\ \hline
		\end{tabular}
\end{table}

Based on the previous experimental results and our analysis of different impacting factors on the
retrieval performances, we propose a new multi-scale image feature representation.
For a given image in the dataset, the whole process of image feature representation is divided into two steps.
First, the input image is fed into the network without the resizing operation (the \textit{free} way) and
a 4-scale feature representation is built on top of the
feature maps of layer conv5\_4. During the multi-scale representation step,
max-pooling of feature maps are used and regional vectors from the same scale are added together and
$l_2$-normalized. After that, features from different scales are summed and $l_2$-normalized again.
 The second step involves applying the PCA and whitening operations on features from the first step.
 The PCA and whitening matrix used are either learned from different or same dataset:
specifically, for the Oxford5k and Oxford105k, it is learned in the Paris6k, while for Paris6k and UKB,
it is learned on Oxford5k and UKB respectively.
The final PCA and whitened image features are used for reporting our method's performances.

\noindent
\textbf{Layer ensemble}. Inspired by previous work on model ensemble to boost the classification
performances~\citep{krizhevsky2012, Simonyan14}, we consider fusing the similarity score from different layers
to improve the retrieval performances. Specifically, for two images, their similarity score is computed as
the weighted sum of the scores from different layers (these weights sum to 1 so that overall similarity score
 between two images are still in the range $[0,1]$.). We have evaluated various combination of layers to see their
performances and find that best performance is achieved by combining the score from conv5\_4 and fc6-conv.
For the fc6-conv features of an image, we use a 3-scale representation as the size of output feature maps
are already very small. The fc6-conv features are compressed to low dimensional vectors for faster computation.
Our layer ensemble achieves 75.6\% and 73.7\% on Oxford5k for the full and cropped queries
respectively, showing a large improvement over previous methods. This suggests that features from the fc6-conv
and conv5\_4 are complementary. See Table~\ref{table:comparsion_with_soa} for the complete results on
all four datasets.

\noindent
\textbf{Comparison}.
We compare the performance of our method with several state-of-the-art methods which use small footprint
representations and do not employ the complicated post-processing techniques such as geometric
re-ranking~\citep{Philbin07} and query expansion~\citep{Relja2012}. The results are shown in
Table~\ref{table:comparsion_with_soa}. In all the datasets and different scenarios (full or cropped),
our method achieves the best performance with comparable cost. For Oxford5k (cropped) and UKB dataset,
the relative improvement of our best results over previous methods (from~\citet{Tolias2016} and
\citet{Babenko2015}) are 10.3\% and 4.4\%.
\section{Conclusion}
In this paper, we focus on instance retrieval based on features extracted from CNNs. we have conducted
extensive experiments to evaluate the impact of five factors on the performances of image retrieval
and analysed their particular impacts.
Based on the insights gained from these experiments, we have proposed a new multi-scale image representation
which shows superior performances over previous methods on four datasets. When combined with the technique
``layer ensemble'', our method can achieve further improvements. Overall, we have provided a viable
and efficient solution to apply CNNs in an unsupervised way to datasets with a relatively small number of
images.

{\small
\bibliography{iclr2017_submit}
\bibliographystyle{iclr2017_conference}
}

\newpage

\begin{appendices}

\section{The network transformations}\label{app:trans}
In order for the network to process images of varying sizes, We change the layer fc6, fc7 and fc8 from
the original model to fc6-conv, fc7-conv and fc8-conv.
It should be noted there are certain constraints on the input image size due
to the network's inherent design.
The original network accepts an image of fixed size ($224\times 224$), so the output feature maps of
the last convolutional layer conv5\_4 is of size $512\times 7 \times 7$. As a result, when we
change the operation between layer conv5\_4 and fc6 from inner product to convolution,
each filter bank kernel between conv5\_4 and fc6-conv has size $7\times 7$.
This in turn means that if we are to extract features from layer fc6-conv and above, the minimum
size of an input image must equal to or be greater than 224. For output feature maps of layer conv5\_4
and below, there are no restrictions on the input image size. During the experiment, when we
are extracting features from layer fc6-conv and above, the minimum size of an image is set to
be 224 if it is less than 224.

\section{The detail of overlap in each scale}\label{app:overlap}
In this paper, the overlaps between different regions occur in the 3 and 4 scale pyramid.
A single region in each scale can be specified as the combination of a slice from the
the width and height of the feature map. If a scale has $N \times N$ regions, then
the number of slices in width and height of the feature map are both $N$.
We use the same set of slices for both the width and height in this experiment.

In 3 scale (see Table~\ref{table:multiscale} (b3)), overlap occurs only in scale 2,
and the slice (in the proportion to the length of feature map width or height:
$\{(0, \frac{2}{3}),(\frac{1}{3}, 1)\}$.
In 4 scale v1 (Table~\ref{table:multiscale} (c1)--(c3)), the slices for scale 2
and 3 are $\{(0, \frac{3}{4}), (\frac{1}{4}, 1)\}$ and $\{(0, \frac{2}{4}), (\frac{1}{4}, \frac{3}{4}),
(\frac{2}{4}, 1) \}$.
In 4 scale v2 (Table~\ref{table:multiscale} (c4)(c5)), the slices for scale 2 and 3
are $\{(0, \frac{3}{5}), (\frac{2}{5}, 1)\}$ and $\{(0, \frac{3}{5}), (\frac{1}{5}, \frac{4}{5}), (\frac{2}{5},1)\}$.
In 4 scale v3 (Table~\ref{table:multiscale} (c6)--(c8)), the slices are $\{(0, \frac{4}{6}), (\frac{2}{6}, 1)\}$
and $\{(0, \frac{3}{6}), (\frac{1}{6}, \frac{4}{6}), (\frac{3}{6}, 1)\}$, for scale 2 and 3, respectively.

\end{appendices}

\end{document}